# Large Pre-Trained Models with Extra-Large Vocabularies: A Contrastive Analysis of Hebrew BERT Models and a New One to Outperform Them All

Eylon Gueta[1,†], Avi Shmidman[1,2,‡], Shaltiel Shmidman[2,‡], Cheyn Shmuel Shmidman[2,†]
Joshua Guedalia[1,2,‡], Moshe Koppel[1,2,‡], Dan Bareket[1,†], Amit Seker[1,†], Reut Tsarfaty[1,3,‡]

[1]Bar Ilan University / Ramat Gan, Israel      [2]DICTA / Jerusalem, Israel
[3]Allen Institute for Artificial Intelligence

[†]guetaeylon@gmail.com, cheynshmuel@gmail.com, dbareket@gmail.com, askere00@gmail.com
[‡]{avi.shmidman,josh.guedalia,shaltiel.shmidman,moshe.koppel,reut.tsarfaty}@biu.ac.il

## Abstract

We present a new pre-trained language model (PLM) for modern Hebrew, termed *AlephBERT-Gimmel*, which employs a much larger vocabulary (128K items) than standard Hebrew PLMs before. We perform a contrastive analysis of this model against all previous Hebrew PLMs (*mBERT, heBERT, AlephBERT*) and assess the effects of larger vocabularies on task performance. Our experiments show that larger vocabularies lead to fewer splits, and that reducing splits is better for model performance, across different tasks. All in all this new model achieves new SOTA on all available Hebrew benchmarks, including Morphological Segmentation, POS Tagging, Full Morphological Analysis, NER, and Sentiment Analysis. Subsequently we advocate for PLMs that are larger not only in terms of number of layers or training data, but also in terms of their vocabulary.[1]

## 1 Introduction

High resource languages such as English enjoy many available pre trained language models (RoBERTa (Liu et al., 2019), BART (Lewis et al., 2020), T5 (Raffel et al., 2020), and more), with a wide range of vocabulary sizes. Many of these models build upon the initial BERT release (Devlin et al., 2019), which tokenizes the source corpora into a finite set of word pieces, limited by the vocabulary size. However, for Hebrew, the choice of vocabulary sizes has heretofore been rather limited. The largest available Hebrew model until now was AlephBERT (Seker et al., 2021), with a vocabulary of 50K tokens; other previous models had Hebrew vocabularies of 32K (heBERT, Chriqui and Yahav (2021)) or 3K (thus Google's multilingual mBERT[2]). In this paper, we introduce a new Hebrew model with a vocabulary of 128K tokens, which provides us with the opportunity, for the first time, to comprehensively examine the effect of different vocabulary sizes upon the overall accuracy of Hebrew BERT models.

Initial studies on BERT with morphologically rich languages (MRLs) argued that this word-pieces architecture was not sufficiently adequate for handling the complications of MRLs (Klein and Tsarfaty, 2020). Some studies overcame this problem by adding a morphological analyzer as part of a preprocess, breaking up units into linguistically relevant morphemes prior to its input into BERT; for instance, in Nzeyimana and Niyongabo Rubungo (2022). Nevertheless, the recent release of AlephBERT (Seker et al., 2021) for Hebrew demonstrated that with a sufficient amount of source data, even the default BERT word piece architecture can succeed in achieving SOTA performance on downstream MRL tasks; further, a recent study demonstrated that character-based models do not necessarily improve task performance for MRLs, above and beyond the standard word-piece paradigm (Keren et al., 2022).

In this paper we shed a new light on this paradigm by assessing the impact of larger vocabulary sizes on pretrained Hebrew BERT models. Our experiments show that, indeed, simply increasing the vocabulary size — without any increase in the size of the source data — consistently increases accuracy on all available Hebrew benchmarks, including morphological segmentation, POS tagging, morphological analysis, named entity recognition, and sentiment analysis. The gains are most pronounced with regard to tasks within the semantic realm. Moreover, we are able to empirically show that an increased number of splits has clear detrimental effects on task performance. All in all, the results in this study indicate that an effective way to boost the accuracy of PLMs is to pre-train with larger vocabularies, in particular regarding

---

[1]We release the new model publicly for unrestricted use: https://github.com/Dicta-Israel-Center-for-Text-Analysis/alephbertgimmel

[2]https://github.com/google-research/bert/blob/master/multilingual.md

| PLM | Hidden Layers | Hebrew Dataset* | Vocabulary |
|---|---|---|---|
| mBERT | 12 | W | 2.4K[5] |
| heBERT | 12 | OW | 30K |
| AlephBERT$_{base}$ | 12 | OWT | 52K |
| ABG$_{small}$ | 6 | OWT | 128K |
| ABG$_{base}$ | 12 | OWT | 128K |

*W - Wikipedia, O - Oscar, T - Tweets

Table 1: Comparing Hebrew-supported PLMs in terms of size (hidden layers), training data and wordpiece-vocabulary size

languages where there is no feasible way to obtain substantially larger amounts of pre-training data.

## 2 A New Model: *AlephBERTGimmel*

The new model we deliver, dubbed *AlephBERT Gimmel* (henceforth *ABG*), is trained on the same dataset as the previous SOTA Hebrew PLM *AlephBERT* (Seker et al., 2021), consisting of approximately 2 billion words of text[3] but with a substantially increased vocabulary, of 128,000 word pieces. Additionally, *ABG* is trained for both MLM and NSP objectives, as opposed to *AlephBERT*, which was trained only for the MLM objective.

We trained both **BERT-small** and **BERT-base** configurations. Both models were trained on a DGX Workstation, containing 4 A100 40GB GPUs, using the NVIDIA optimized extensions to the HuggingFace library. (Wolf et al., 2019)[4] For the **BERT-small** model we used a a batch size of 12,288, which maximized the 40GB memory capacity of the GPUs. The model trained for 13,200 batches with an initial learning rate of 6e-3. Total training time was 3 days. For the **BERT-base** model we used a smaller batch size of 8192, in order to fit the training within the GPU memory. The model first trained for 31,400 batches with an initial learning rate of 6e-3 and we continued training for an additional 37,800 batches with an initial learning rate of 1e 4. Total training time was 18 days.

## 3 Experimental Setup

**Models** We compare the performance of *ABG* with all existing Hebrew BERT instantiations. The differences between these PLMs, for all tasks we experimented with, are summarized in Table 1.

---

[3]Specifically, the model was trained on **Oscar**, **Twitter**, and **Wikipedia**; for more specifics regarding each of these corpora, see Seker et al. (2021)

[4]https://github.com/NVIDIA/DeepLearningExamples/tree/master/PyTorch/LanguageModeling/BERT

[5]For mBERT we only consider the Hebrew portion of the multi-lingual vocabulary

**Tasks** Modern Hebrew is a Semitic language with rich morphology and complex orthography. As a result, the basic processing units in the language are typically smaller the span of raw words. Consequently, most standard evaluation tasks require knowledge of internal morphological boundaries within the raw words.

To accommodate the granularity of these evaluation tasks, as part of our PLM processing pipeline we extract morphological segments for each word and label each segment using the morphological extraction and labeling component of Seker et al. (2021). The input to each of these evaluation tasks is a sequence of raw tokens and the output is a morphological-level sequence.[6] Table 2 provides a description for all relevant tasks.

To gauge the effect of segmentation on downstream tasks we define an additional morpheme-level NER task where we replace tokens with their corresponding gold segments as the input sequence. Additionally, in order to isolate the effect of the quality of the contextualized vectors produced by the PLMs on the downstream task we follow Seker et al. (2021) and experiment with word-level NER predictions - achieved by directly fine-tuning the PLMs using an additional token-classification head.[7] Finally we predict sentence level class labels by directly fine-tuning the PLMs using an additional sentence-classification head, Assign one of Positive, Negative, Neutral to the entire input sequence.

We follow the tasks setups used by (Seker et al., 2021): For both Seg & POS & feats and Seg & NER we use the multi-task configuration of (Brusilovsky and Tsarfaty, 2022; Seker et al., 2021). Each of the tasks is trained separately resulting in 2 models. For NER (tokens/gold Seg) we use huggingface/transformers (Wolf et al., 2019) token-classification, and for Sentiment sequence-classification.

**Datasets** We set a new standard, unifying the previous used UD (Sade et al., 2018), SPMRL (Seddah et al., 2013) and NEMO (Bareket and Tsarfaty, 2021), as these corpora use the same texts but previous annotations do not fully align. We then train and test all models on the a fixed version we created, of UD joined with NEMO as a stand-off NER annotation, and which we henceforth nickname *UD-NEMO*. Both Seg & POS & feats, Seg & NER and NER (token/morph) are performed against this dataset. The sentiment task is performed against its own dataset (Amram et al., 2018).

---

[6]These units comply with the 2 level representation of tokens defined by UD, each unit with a single POS tag. https://universaldependencies.org/u/overview/tokenization.html

[7]In cases where a word is split into multiple wordpieces by the PLM tokenizer, we employ common practice and use the first wordpiece vector.

| Task | Input | Output |
|---|---|---|
| Segmentation | A sequence of raw tokens | a sequence of morphological segments |
| POS Tagging | A sequence of raw tokens | a sequence of morphological segments and their associated POS tags |
| Morph Analysis | A sequence of raw tokens | a sequence of morphological segments and their associated morphological features |
| Morph. NER | A sequence of raw tokens | a sequence of morphological segments and their associated BIOSE + entity type |
| Token NER | A sequence of raw tokens | the input sequence of tokens along with a single BIOSE + entity type |
| Sentiment | A sentence | Positive, Negative, Neutral. |

Table 2: Hebrew Tasks for PLM Assessment

| | Seg & POS & feats | | |
|---|---|---|---|
| | Seg (F1) | POS (F1) | Features (F1) |
| mBERT | 96.07 | 93.14 | 92.68 |
| heBERT | 97.90 | 95.80 | 95 35 |
| AlephBERT | 97.88 | 95.81 | 95.27 |
| ABG-small | 97 27 | 94.79 | 94.38 |
| ABG-base | **98.09** | **96.22** | **95.76** |

Table 3: Morpheme-Based Aligned MultiSet (mset) results on the UD-NEMO Corpus.

| | Sentiment |
|---|---|
| | Accuracy |
| mBERT | 84.21 |
| heBERT | 87 13 |
| AlephBERT | 89.02 |
| ABG-small | 89.00 |
| ABG-base | **89.51** |

Table 5: Sentiment Analysis Scores on the Facebook Corpus. Previous SOTA is by (Seker et al., 2021)

| | Seg & NER MTL | | NER Token | NER Seg |
|---|---|---|---|---|
| | Seg F1 | NER F1 | NER F1 | NER F1 |
| mBERT | 96.66 | 69.78 | 79.11 | 77.01 |
| heBERT | 98.04 | 77.29 | 81.13 | 82.66 |
| AlephBERT | 97.84 | 77.55 | 83.62 | 83.43 |
| ABG-small | 97.43 | 73.86 | 78.88 | 77.02 |
| ABG-base | **98.22** | **80.39** | **86.26** | **85.26** |

Table 4: NER Evaluation on the UD-NEMO Corpus.

**Metrics** We follow the evaluation used by (Seker et al., 2021). For Seg, Seg & POS. Seg & feats, we report the mset score. For Seg & NER we report the mset score for Seg, and NEMO F1 score for NER. For NER we report the NEMO F1 score. For Sentiment, we report sentence accuracy.[8] All results are reported on the standard test set.

## 4 Results and Analysis

Tables 3, 4, 5 show the results of all models on all tasks. As we can see, ABG-base sets a new SOTA for all reported tasks.

For the morphological tasks (Table 3), Aleph-BERT and heBERT alternately achieve the 2nd best result, with a very small gap from ABG. For

[8]See a discussion of the metrics in Seker et al. (2021).

NER, on all of its variants (Table 4), ABG achieves a significant improvement over all other models. It is interesting to see that ABG-small achieves near-SOTA results for both Segmentation and Sentiment (Table 5) tasks. This might suggest both an insight into the difficulty of these tasks, and a practically competitive small model, having half the amount of layers, but equipped with a large vocabulary.

**Word-pieces analyses** In order to examine the impact of models' vocabulary sizes on the models' performance, we first examine the tasks' corpus with respect to the vocabulary size. Figure 2 shows that, as expected, with the increase of vocabulary size more words of the tasks corpus are available to the model as a single piece, and less split tokens are obtained on the whole.

When we examine the impact of the number of pieces (henceforth, *#pieces*) per token on the models performance, we compare tokens with respect to split *#pieces*: tokens (or spans in NER) are grouped according to their *#pieces*, and evaluation is performed on each of the following groups: 1 (token is in vocabulary), 2 (token is split to two pieces), or 3+ pieces (token is split to 3 or more *#pieces*). For NER, as this is a span-level task rather then token level, we take a span-wise approach: mentions which are spans of segments, are treated as a single unit (Fig. 3). These spans are evaluated against their parallel tokens predictions and grouped according to a binary version of *#pieces*: whether at least one of their tokens is split by the tokenizer or not. For each group we report its NEMO F1 scores.

Figure 1 presents the impact of *#pieces* on Seg

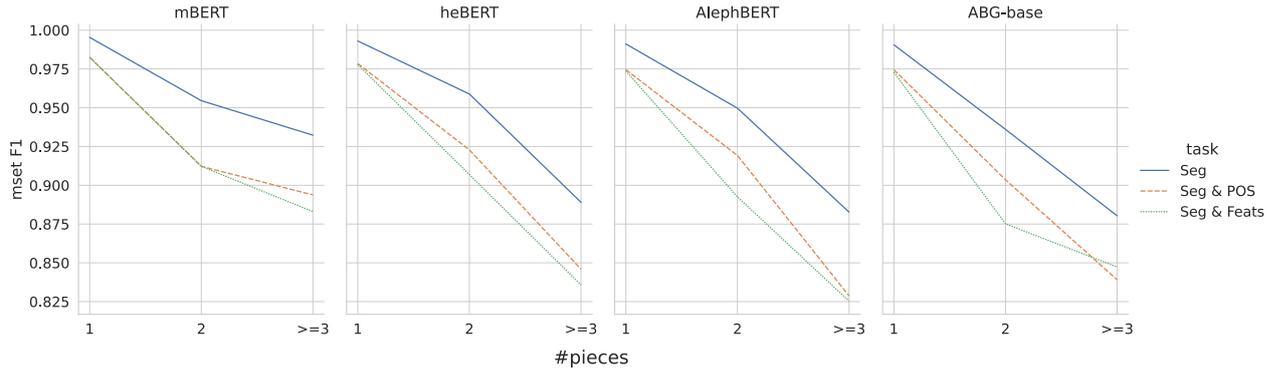

Figure 1: F1 (mset) scores by #pieces per token for each PLM for Seg, POS and morphological features (test set).

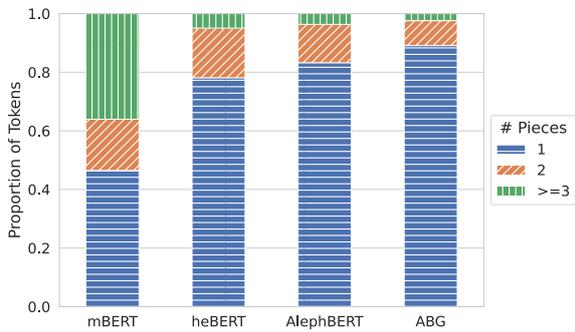

Figure 2: Distribution of *#pieces* per-token for each PLM (test set).

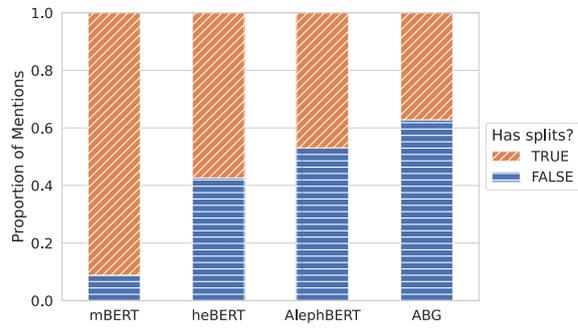

Figure 3: Distribution of NER mentions with/without wordpiece splits for each PLM (test set).

& POS & feats. All models perform near-perfect when the token is in the vocabulary (#pieces= 1), but suffer as #pieces increase. This break down allows to observe the supposedly high results achieved by all models, exposing their true weakness when it comes to OOV tokens with respect to the underlying tokenizer.

As for NER, Figure 4 presents an even more pronounced impact of #pieces on performance; all models show a sharp decrease when even a single token within the mention is split. All in all, we conclude that increasing the size of the word-pieces vocabulary can be an effective way to boost PLMs performance, in particular when additional pre-train data is not easily attainable, as it is in low-to-medium resourced languages such as Hebrew.

## 5 Conclusion

In this paper we present AlephBERTGimmel, a BERT-based Hebrew PLM, with a vocabulary size far larger than all Hebrew PLMs before. The new model performs higher on all modern Hebrew benchmarks, setting a new SOTA for modern He-brew NLP. Beyond that, our results clearly demonstrate the critical importance of choosing a large enough vocabulary size for pre-training. To be sure, there is a tradeoff to be considered between vocabulary size and training time; larger vocabularies require more GPU memory, thus a smaller batch size, and for the same amount of clocktime a higher vocabulary will mean fewer epochs over the data. Nevertheless, the palpable gains afforded by the higher vocabulary may well outshine the incremental gains provided by the extra epochs.

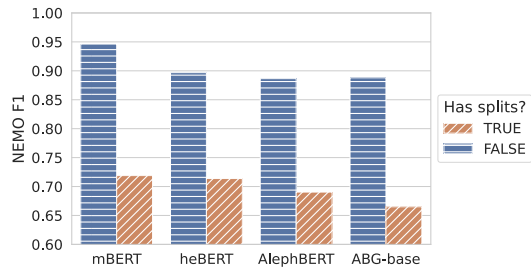

Figure 4: Seg & NER NEMO F1 scores for mentions with/without wordpiece splits for each PLM (test set).

# 6 Limitations

The primary limitation of the AlephBERTGimmel model described here is the relatively narrow range of genres contained in its source corpora. Due to the lack of large public-domain textual corpora available for Modern Hebrew, we had to rely upon Wikipedia, Twitter, and the OSCAR common crawl. Thus, other genres are unrepresented or underrepresented, including scientific writing and *belles-lettres*. Furthermore, because of the heavy presence of the twitter corpus, the model has internalized the use of many terms considered slang or informal. In and of itself, this is an advantage rather than a weakness, because it means that our model reflects the language as it is used in practice, and that it will be able to contend with texts of lower register often found on social media. However, if our model were to be deployed for word prediction, certain contexts may results in inappropriate word suggestions, and these may need to be filtered out in post-processing, as relevant.